\documentclass[manuscript, nonacm]{acmart}

\copyrightyear{2022}
\acmYear{2022}

\acmConference[RecSys '22]{Sixteenth ACM Conference on Recommender Systems}{ September 18 -- 23, 2022}{Seattle, WA, USA}
\acmBooktitle{Sixteenth ACM Conference on Recommender Systems (RecSys '22),  September 18 -- 23, 2022, Seattle, WA, USA}

\acmDOI{}
\acmISBN{}

\AtBeginDocument{%
  \providecommand\BibTeX{{%
    \normalfont B\kern-0.5em{\scshape i\kern-0.25em b}\kern-0.8em\TeX}}}

\usepackage{lipsum}
\usepackage{float}
\usepackage{soul}
\usepackage{color}
\usepackage{bbm}
\usepackage{enumitem}
\usepackage{caption}
\usepackage{subcaption}
\usepackage[toc,page]{appendix}

\definecolor{Blue}{rgb}{0,0,1}

\definecolor{Orange}{rgb}{1,0.5,0}

\definecolor{Green}{rgb}{0,1,0}

\begin{document}
\fancyhead{}
\pagenumbering{gobble}

\title{Extending Open Bandit Pipeline to Simulate Industry Challenges}

\author{Bram van den Akker}
\email{bram.vandenakker@booking.com}

\author{Niklas Weber}
\email{niklas.weber@booking.com}

\author{Felipe Moraes}
\email{felipe.moraes@booking.com}

\author{Dmitri Goldenberg}
\email{dima.goldenberg@booking.com}

\affiliation{\institution{Booking.com}
\city{Amsterdam}
\country{The Netherlands}}

\begin{abstract} 
Bandit algorithms are often used in the e-commerce industry to train Machine Learning (ML) systems when pre-labeled data is unavailable. However, the industry setting poses various challenges that make implementing bandit algorithms in practice non-trivial. In this paper, we elaborate on the challenges of off-policy optimisation, delayed reward, concept drift, reward design, and business rules constraints that practitioners at Booking.com encounter when applying bandit algorithms. 
Our main contributions is an extension to the Open Bandit Pipeline (OBP) framework. We provide simulation components for some of the above-mentioned challenges to provide future practitioners, researchers, and educators with a resource to address challenges encountered in the e-commerce industry.

\end{abstract}



\maketitle

\section{Introduction} \label{sec:introduction}

In recent years the application of bandit algorithms in e-commerce has yielded convincing results~\cite{liu2021map}. These methods are commonly used when pre-labeled data is non-trivial to obtain or unavailable. 
At Booking.com we have experienced many challenges while designing production ML systems using bandit algorithms, which have been described extensively in previous work~\cite{goldenberg2021know, booking2021personalization}. These challenges include: \textbf{(i)} choosing between on-policy and off-policy algorithms; \textbf{(ii)} deciding on how to handle delayed reward; \textbf{(iii)} dealing with various types of concept drift; \textbf{(iv)} defining realistic rewards; and \textbf{(iv)} optimising bandit policies under external business rules with constantly changing available arms. 

A critical consequence of overlooking the challenges mentioned above is that it could cause practitioners to end up with sub-optimal policies. Ignoring these challenges may slow down their work, decrease incremental revenue, and create barriers to understanding why the bandit algorithms do not work in practice.

Research on bandit algorithms frequently consists of executing experiments on open source data, proprietary data, and simulations. Similarly, practitioners in industry need simulators to speed up experimentation \cite{bernardi2021simulations}.
Various simulation frameworks for recommender systems and bandit algorithms exist \cite{rohde2018recogym, ie2019recsim, zuo2021recsys, shi2019virtual, huang2020keeping}. Unfortunately, the majority of these simulations have limited documentation,  tests, and ongoing development activity. These limitations make it challenging to use and extend these frameworks. 
Recently, \citet{saito2020open} released the Open Bandit Pipeline (OBP), which provides a modular and extendable framework for researching bandit algorithms and off-policy methods. To the best of our knowledge, this simulation framework has the most extensive documentation, code quality standards, and extensibility of the available open-source bandit simulation frameworks.
In this paper, we extend OBP by providing new components\footnote{https://github.com/st-tech/zr-obp/pull/177} that can facilitate researchers and educators in the field of bandit algorithms to start addressing the industry challenges we experienced at Booking.com. In particular, we aim to provide a resource that future practitioners can refer to as they address the challenges we encounter in the e-commerce industry.

\section{Industry Bandit Challenges}~\label{sec:challenges}
In this section we expand on the industry challenges described above and introduce some key questions that practitioners encounter when trying to address them.
\subsection{On-policy vs Off-policy} \label{sec:onoffpolicy}

Bandit algorithms are commonly optimised on-policy. This means that at each round, an action is sampled to either exploit from the policy or explore. Subsequently, the production policy is updated with the observed reward in that round. Off-policy optimisation, in contrast, uses already-existing training data to learn a policy, which is then deployed to production. This is commonly achieved by optimising a policy using of some form of inverse-propensity weighting (IPW) to debias the logged data\cite{strehl2010learning}. To the best of our knowledge there is limited work providing insights under which circumstances on-policy or off-policy methods trained on logged data are more favorable. Some studies provide recommendations on using on-policy and off-policy in settings such as learn to rank ~\cite{jagerman2019model}. Those commonly favor the use of on-policy learning or propose to combine both on-policy and off-policy~\cite{oosterhuis2021unifying, o2016pgq}. 
In practice, on-policy learning can be expensive in terms of experimentation and training as each policy needs to be able to interact with users, possibly negatively affecting user experience. Similarly, when comparing multiple candidate policies, each policy has to be able to explore independently. By using off-policy methods practitioners can compare endless configurations in parallel without affecting the user experience. 
In practice, we are often faced with the following challenges when deciding between on-policy and off-policy methods:

\begin{enumerate}[label=\textbf{Q\ref{sec:onoffpolicy}.\arabic*}]   
    \item \textit{In which scenarios does off-policy learning outperform on-policy methods?}\label{scenariooffpolicy}
    \item \textit{How frequently should off-policy trained models be updated?}
    \item \textit{Which strategies of collecting counterfactual examples for off-policy (re-)training are most effective?}\label{strategycounterfactual}
\end{enumerate}


\subsection{Concept Drift \& Non-stationarity} \label{sec:conceptdrift}

Productionisation of ML systems often faces the challenge of learning under concept drift~\cite{lu2018learning,vzliobaite2016overview, tax2021machine, ktena2019addressing}, where feature and label distributions are non-stationary and evolve over time. For instance, at Booking.com, we observed users' preferences changing during impactful events, such as the Covid pandemic. During this time preferences shifted to domestic trips due to restrictions on traveling abroad~\cite{goldenberg2021know}. 
The events affecting the environment can cause different types of drift~\cite{cavenaghi2021non} such as: (i) sudden/abrupt, like natural disasters; (ii) incremental events, during slow-moving events such as economic crises; (iii) gradual, during events such as the Covid pandemic; and (iv) recurring/seasonal, during events such as summer vacations or Christmas breaks. For example, seasonality is a kind of concept drift that typically happens in industry settings such as in e-commerce, as well in other domains such as in music streaming platforms. Here people's music taste changing over time as shown by~\cite{di2020linear}. 
While the classic stochastic bandit assumes a stationary environment\cite{lattimore2020bandit}, various bandit algorithms have been proposed that adapt in non-stationary environments. ~\cite{slivkins2008adapting, zeng2016online, luo2018efficient, hong2021non, saha2022non-stationary}. 
However, in practice we often face questions such as:

\begin{enumerate}[label=\textbf{Q\ref{sec:conceptdrift}.\arabic*}]    \item \textit{How many rounds does it take for on-policy methods to recover from drift?}
    \item \textit{Which types of drift are more challenging in different settings and for different algorithms?}\label{question:drift}
    \item \textit{How to apply off-policy methods when your environment is non-stationary?}
\end{enumerate}

\subsection{Delayed Reward} \label{sec:delayedfeedback}

In e-commerce, we observe that reward may take many rounds to arrive, as opposed to the theoretical bandit formulation in which the reward is typically available after each prediction~\cite{lattimore2020bandit, tsagkias2021challenges}. The length of the delay can depend on various factors. For instance, in marketing e-mails we expect a shorter delay from clicks on e-mails and a longer delay from purchases after opening an e-mail. In the simplest case a delay is unbiased, eg. constant or sampled from a distribution. In more complicated cases the delay distribution can depend on the context, action, or even the expected reward of an arm, e.g.\ if a less-desirable recommendation causes a user to hesitate longer.
The authors of \cite{joulani2013online} showed that delay causes additive regret in stochastic bandits, while being multiplicative in adversarial problems. Recently, many approaches to learning under delayed reward have been proposed~\cite{ktena2019addressing, lancewicki2021stochastic, thune2019nonstochastic, thune2019nonstochastic, bernardi2020recommending}. However, in practice we commonly face uncertainty about the effects of delays on model performance, such as:

\begin{enumerate}[label=\textbf{Q\ref{sec:delayedfeedback}.\arabic*}]
    \item \textit{What is the extra regret due to delayed reward in (non-)stationary environments?}\label{question:delaydrift}
    \item \textit{Is the delay biased and how does this influence the policy performance? }\label{question:delaybias}
\end{enumerate}

\subsection{Reward Design and Multi-Reward}\label{sec:rewarddesign}
In an ideal scenario the reward for a bandit algorithm is identical to its business objective~\cite{dragone2019deriving}. However, in practice directly using a business objective, e.g.\ conversions, as rewards is commonly infeasible due to sparsity, delay, or attribution complexity. For example, we could use a bandit algorithm to decide whether a user should be incentivized to book a hotel with either 1) a free taxi with every booking, or 2) 15\% discount on every booking, and show it as a banner in the search results. We would like to reward the model if the banner leads to a booking. However, determining whether a booking was made due to the picked banner, and not any other part of the search results page, is non-trivial. On the other hand, determining whether a given banner was seen or clicked can be straightforward, but is a weaker indicator of commercial success. 
Even when a conversion can be attributed to an action, this reward can still be too sparse or too delayed for the algorithm to learn effectively. In those cases, practitioners can decide to use a surrogate reward such as clicks \cite{chagniot2020clicks} or combining multiple rewards within a funnel~\cite{liu2021funnel, 10.1145/3394486.3403374} to speed up the learning process. However, these types of surrogate rewards might lead to sub-optimal policies or even clickbait. 
To effectively design rewards for bandit algorithms in practice, we are commonly faced with questions such as: 

\begin{enumerate}[label=\textbf{Q\ref{sec:rewarddesign}.\arabic*}]
    \item \textit{What is the effect of using a surrogate reward on your primary business metric?}
    \item \textit{How do we deal with sparse primary rewards?}
    \item \textit{How can we optimise a bandit model when attribution of rewards to actions is challenging?}
    \item \textit{Can we combine multiple rewards effectively to overcome challenges in our primary reward?}
\end{enumerate}

\subsection{Business Rules and Arm Availability}\label{sec:businessrules}
In literature, the bandit problem is often presented in isolation. A policy samples actions, such as travel destinations based on a visitors context, and observes the reward of this action. In practice, these actions are typically subject to various business constraints. For example, only destinations with available rooms can be shown, or similar destinations should not be shown in the same set of recommendations. Consequently, actions sampled by the policy may not be shown, which restricts the ability of the policy to control exploration. Additionally, debiasing the logged data is non-trivial, as the propensities of some actions can become  very small or even non-existent at all. 
Recent work has proposed to factor out business rules in a stand-alone components~\cite{10.1145/3447548.3467165, falk2019practical}. However, evaluating the legality of all possible actions before calling the model can be non-trivial or expensive. Previous research has yielded bandit variants incorporating various constraints on arm availability directly into the bandit model, e.g. (Contextual) Blocking Bandits \cite{50090, Basu2019}, Sleeping Bandits \cite{kleinberg2010regret}, Any-m Feasible Arm Identification \cite{Bagherjeiran2019}, or, in some sense, Bayesian Meta-Learning \cite{Nabi2021}.

In addition to business rules adjusting the available actions on each round, the pool of available arms is likely to change over the bandit's life-cycle. Many existing algorithms naturally extend to introducing new arms in principle but suffer from it in practice~\cite{liu2018incentivizing, ghalme2021ballooning}. \cite{liu2018incentivizing}, show UCB suffering linear regret in a scenario in which arms get added frequently, leading to over-exploration. Questions arising in practice are: 

\begin{enumerate}[label=\textbf{Q\ref{sec:businessrules}.\arabic*}]
    \item \textit{How is the performance of various bandit models affected when sampled arms are blocked by the system?}
    \item \textit{How do we debias data produced with biased business rules?}
    \item \textit{How can off-policy models be used when arms are continuously added and removed?}
    \item \textit{How can we effectively learn policies from a large pool of constantly changing arms?}
\end{enumerate}

\section{Extensions to Open Bandit Pipeline} \label{sec:simulation}
In this section we will expand on the components we introduced in OBP to facilitate researchers and educators with simulations that represent the industry challenges we experience at Booking.com. We restrict ourselves to the concepts described in sections~\ref{sec:onoffpolicy},~\ref{sec:conceptdrift}, and~\ref{sec:delayedfeedback}. The remaining challenges are left for future work.

We extend OBP with a \textit{BanditEnvironmentSimulator} and \textit{BanditPolicySimulator} class. The \textit{BanditEnvironmentSimulator} samples one or multiple rounds of contexts and rewards for all arms. The \textit{BanditPolicySimulator} takes any bandit algorithm and iterates over the generated rounds to select actions and update the policy.  

To allow experiments that can answer the question in Section~\ref{sec:onoffpolicy}, the \textit{BanditPolicySimulator} keeps a log of its behaviour. This allows us to use these logs in the off-policy parts of the OBP framework. Using the reward of all arms provided by the \textit{BanditEnvironmentSimulator} we can compare the performance of on-policy and off-policy methods on the same dataset. An example of such a simulation can be found in Appendix~\ref{sec:offpolicyappendix}.

For the challenges described in Section~\ref{sec:conceptdrift}, we implemented functionality to simulate various types of drift. The simulations calculate an expected reward based on the interaction of a user-context, action-context, and a set of matching coefficients. We introduce a \textit{CoefficientDrifter} class, which controls the coefficients in each round based on an \textit{interval}, \textit{transition period}, \textit{transition type}, and \textit{seasonality} parameter. 
These parameters can be used to create the four types of concept drift as described by~\cite{cavenaghi2021non} and more. An example using the \textit{CoefficientDrifter}, showing how sudden and seasonal drift affect the performance of existing bandit algorithms, can be found in Appendix~\ref{sec:offpolicyappendix}.

Finally, for the challenges described in Section~\ref{sec:delayedfeedback}, we extended the \textit{BanditEnvironmentSimulator} with a \textit{delay\_function} parameter. When provided, this function adds a \textit{delay\_rounds} field to each round. If this field is present, the model parameters are only updated after \textit{delay\_rounds} rounds have passed, instead of updating immediately after each round.

Additionally, we provide an \textit{ExponentialDelaySampler} containing both an unbiased and \emph{reward-dependent}\cite{lancewicki2021stochastic} delay function. In unbiased delay, $d_t$ is sampled from an exponential distribution. In this scenario, the delay is the same for each arm per round. However, in the  \emph{reward-dependent}~\cite{lancewicki2021stochastic} setting the random delay $d_t$ at each round depends on the individual expected rewards for each arm in that round. The sample is taken by interpolating between two exponential distributions at different scales. 
We use exponential distributions for delays as they are commonly found in real-world situations as described by~\cite{chapelle2014modeling, ktena2019addressing}. In Appendix~\ref{sec:rewardappendix} you can find an experiment demonstrating how the delay functions can be used in a stationary environment.

\section{Conclusion and Future Work} \label{sec:conclusion}

In this paper, we describe challenges we encounter in the industry and provide ways to simulate those challenges by extending OBP. Our extended version and working notebooks are available for future practitioners interested in answering questions around the challenges we elaborated on. We hope our contributions can speed up the work of industry practitioners as well as researchers and serve as a tool for educational purposes.
In future work we want to continue extending OBP additional simulation capabilities covering more of our described challenges as well as implement functionality to tackle other challenges such as slate recommendations.

\pagebreak
\bibliographystyle{ACM-Reference-Format}
\bibliography{main}


\begin{thebibliography}{45}


\ifx \showCODEN    \undefined \def \showCODEN     #1{\unskip}     \fi
\ifx \showDOI      \undefined \def \showDOI       #1{#1}\fi
\ifx \showISBNx    \undefined \def \showISBNx     #1{\unskip}     \fi
\ifx \showISBNxiii \undefined \def \showISBNxiii  #1{\unskip}     \fi
\ifx \showISSN     \undefined \def \showISSN      #1{\unskip}     \fi
\ifx \showLCCN     \undefined \def \showLCCN      #1{\unskip}     \fi
\ifx \shownote     \undefined \def \shownote      #1{#1}          \fi
\ifx \showarticletitle \undefined \def \showarticletitle #1{#1}   \fi
\ifx \showURL      \undefined \def \showURL       {\relax}        \fi
\providecommand\bibfield[2]{#2}
\providecommand\bibinfo[2]{#2}
\providecommand\natexlab[1]{#1}
\providecommand\showeprint[2][]{arXiv:#2}

\bibitem[\protect\citeauthoryear{Bagherjeiran and Katz-Samuels}{Bagherjeiran
  and Katz-Samuels}{2019}]%
        {Bagherjeiran2019}
\bibfield{author}{\bibinfo{person}{Abraham Bagherjeiran} {and}
  \bibinfo{person}{Julian Katz-Samuels}.} \bibinfo{year}{2019}\natexlab{}.
\newblock \showarticletitle{Any-m feasible arm identification}. In
  \bibinfo{booktitle}{\emph{KDD AdKDD 2019}}.
\newblock
\urldef\tempurl%
\url{https://www.amazon.science/publications/any-m-feasible-arm-identification}
\showURL{%
\tempurl}


\bibitem[\protect\citeauthoryear{Basu, Sanghavi, Sen, and Shakkottai}{Basu
  et~al\mbox{.}}{2019}]%
        {Basu2019}
\bibfield{author}{\bibinfo{person}{Soumya Basu}, \bibinfo{person}{Sujay
  Sanghavi}, \bibinfo{person}{Rajat Sen}, {and} \bibinfo{person}{Sanjay
  Shakkottai}.} \bibinfo{year}{2019}\natexlab{}.
\newblock \showarticletitle{Blocking bandits}. In
  \bibinfo{booktitle}{\emph{NeurIPS 2019}}.
\newblock
\urldef\tempurl%
\url{https://www.amazon.science/publications/blocking-bandits}
\showURL{%
\tempurl}


\bibitem[\protect\citeauthoryear{Benedetto, Bellini, and Zappella}{Benedetto
  et~al\mbox{.}}{2020}]%
        {di2020linear}
\bibfield{author}{\bibinfo{person}{Giuseppe~Di Benedetto},
  \bibinfo{person}{Vito Bellini}, {and} \bibinfo{person}{Giovanni Zappella}.}
  \bibinfo{year}{2020}\natexlab{}.
\newblock \showarticletitle{A linear bandit for seasonal environments}.
\newblock  (\bibinfo{year}{2020}).
\newblock
\urldef\tempurl%
\url{https://www.amazon.science/publications/a-linear-bandit-for-seasonal-environments}
\showURL{%
\tempurl}


\bibitem[\protect\citeauthoryear{Bernardi, Batra, and Bruscantini}{Bernardi
  et~al\mbox{.}}{2021}]%
        {bernardi2021simulations}
\bibfield{author}{\bibinfo{person}{Lucas Bernardi}, \bibinfo{person}{Sakshi
  Batra}, {and} \bibinfo{person}{Cintia~Alicia Bruscantini}.}
  \bibinfo{year}{2021}\natexlab{}.
\newblock \showarticletitle{Simulations in Recommender Systems: An industry
  perspective}.
\newblock \bibinfo{journal}{\emph{arXiv preprint arXiv:2109.06723}}
  (\bibinfo{year}{2021}).
\newblock


\bibitem[\protect\citeauthoryear{Bernardi, Estevez, Eidis, and Osama}{Bernardi
  et~al\mbox{.}}{2020}]%
        {bernardi2020recommending}
\bibfield{author}{\bibinfo{person}{Lucas Bernardi}, \bibinfo{person}{Pablo
  Estevez}, \bibinfo{person}{Matias Eidis}, {and} \bibinfo{person}{Eqbal
  Osama}.} \bibinfo{year}{2020}\natexlab{}.
\newblock \showarticletitle{Recommending Accommodation Filters with Online
  Learning}. In \bibinfo{booktitle}{\emph{Proceedings of the Workshop on Online
  Recommender Systems and User Modeling (ORSUM @ RecSys 2021)}}.
\newblock


\bibitem[\protect\citeauthoryear{Caramanis, Papadigenopoulas, Shakkottai, and
  Basu}{Caramanis et~al\mbox{.}}{2021}]%
        {50090}
\bibfield{author}{\bibinfo{person}{Constantine Caramanis},
  \bibinfo{person}{Orestis Papadigenopoulas}, \bibinfo{person}{Sanjay
  Shakkottai}, {and} \bibinfo{person}{Soumya Basu}.}
  \bibinfo{year}{2021}\natexlab{}.
\newblock \showarticletitle{Contextual Blocking Bandits}. In
  \bibinfo{booktitle}{\emph{International Conference on Artificial Intelligence
  and Statistics}}.
\newblock


\bibitem[\protect\citeauthoryear{Cavenaghi, Sottocornola, Stella, and
  Zanker}{Cavenaghi et~al\mbox{.}}{2021}]%
        {cavenaghi2021non}
\bibfield{author}{\bibinfo{person}{Emanuele Cavenaghi},
  \bibinfo{person}{Gabriele Sottocornola}, \bibinfo{person}{Fabio Stella},
  {and} \bibinfo{person}{Markus Zanker}.} \bibinfo{year}{2021}\natexlab{}.
\newblock \showarticletitle{Non stationary multi-armed bandit: Empirical
  evaluation of a new concept drift-aware algorithm}.
\newblock \bibinfo{journal}{\emph{Entropy}} \bibinfo{volume}{23},
  \bibinfo{number}{3} (\bibinfo{year}{2021}), \bibinfo{pages}{380}.
\newblock


\bibitem[\protect\citeauthoryear{Chagniot, Vasile, and Rohde}{Chagniot
  et~al\mbox{.}}{2020}]%
        {chagniot2020clicks}
\bibfield{author}{\bibinfo{person}{Philom{\`e}ne Chagniot},
  \bibinfo{person}{Flavian Vasile}, {and} \bibinfo{person}{David Rohde}.}
  \bibinfo{year}{2020}\natexlab{}.
\newblock \showarticletitle{From Clicks to Conversions: Recommendation for
  long-term reward}.
\newblock \bibinfo{journal}{\emph{arXiv preprint arXiv:2009.00497}}
  (\bibinfo{year}{2020}).
\newblock


\bibitem[\protect\citeauthoryear{Chapelle}{Chapelle}{2014}]%
        {chapelle2014modeling}
\bibfield{author}{\bibinfo{person}{Olivier Chapelle}.}
  \bibinfo{year}{2014}\natexlab{}.
\newblock \showarticletitle{Modeling delayed feedback in display advertising}.
  In \bibinfo{booktitle}{\emph{Proceedings of the 20th ACM SIGKDD international
  conference on Knowledge discovery and data mining}}.
  \bibinfo{pages}{1097--1105}.
\newblock


\bibitem[\protect\citeauthoryear{Dragone, Mehrotra, and Lalmas}{Dragone
  et~al\mbox{.}}{2019}]%
        {dragone2019deriving}
\bibfield{author}{\bibinfo{person}{Paolo Dragone}, \bibinfo{person}{Rishabh
  Mehrotra}, {and} \bibinfo{person}{Mounia Lalmas}.}
  \bibinfo{year}{2019}\natexlab{}.
\newblock \showarticletitle{Deriving user-and content-specific rewards for
  contextual bandits}. In \bibinfo{booktitle}{\emph{The World Wide Web
  Conference}}. \bibinfo{pages}{2680--2686}.
\newblock


\bibitem[\protect\citeauthoryear{Falk}{Falk}{2019}]%
        {falk2019practical}
\bibfield{author}{\bibinfo{person}{Kim Falk}.} \bibinfo{year}{2019}\natexlab{}.
\newblock \bibinfo{booktitle}{\emph{Practical recommender systems}}.
\newblock \bibinfo{publisher}{Simon and Schuster}.
\newblock
\urldef\tempurl%
\url{https://livebook.manning.com/book/practical-recommender-systems/chapter-6/83}
\showURL{%
\tempurl}


\bibitem[\protect\citeauthoryear{Ghalme, Dhamal, Jain, Gujar, and
  Narahari}{Ghalme et~al\mbox{.}}{2021}]%
        {ghalme2021ballooning}
\bibfield{author}{\bibinfo{person}{Ganesh Ghalme}, \bibinfo{person}{Swapnil
  Dhamal}, \bibinfo{person}{Shweta Jain}, \bibinfo{person}{Sujit Gujar}, {and}
  \bibinfo{person}{Y Narahari}.} \bibinfo{year}{2021}\natexlab{}.
\newblock \showarticletitle{Ballooning multi-armed bandits}.
\newblock \bibinfo{journal}{\emph{Artificial Intelligence}}
  \bibinfo{volume}{296} (\bibinfo{year}{2021}), \bibinfo{pages}{103485}.
\newblock


\bibitem[\protect\citeauthoryear{Goldenberg, Kofman, Albert, Mizrachi,
  Horowitz, and Teinemaa}{Goldenberg et~al\mbox{.}}{2021a}]%
        {booking2021personalization}
\bibfield{author}{\bibinfo{person}{Dmitri Goldenberg}, \bibinfo{person}{Kostia
  Kofman}, \bibinfo{person}{Javier Albert}, \bibinfo{person}{Sarai Mizrachi},
  \bibinfo{person}{Adam Horowitz}, {and} \bibinfo{person}{Irene Teinemaa}.}
  \bibinfo{year}{2021}\natexlab{a}.
\newblock \showarticletitle{Personalization in Practice: Methods and
  Applications}. In \bibinfo{booktitle}{\emph{Proceedings of the 14th
  International Conference on Web Search and Data Mining}}.
\newblock


\bibitem[\protect\citeauthoryear{Goldenberg, Mizrachi, Horowitz, Kangas,
  Schwoerer, Mozzato, Ferretti, Panagiotis, and Bernardi}{Goldenberg
  et~al\mbox{.}}{2021b}]%
        {goldenberg2021know}
\bibfield{author}{\bibinfo{person}{Dmitri Goldenberg}, \bibinfo{person}{Sarai
  Mizrachi}, \bibinfo{person}{Adam Horowitz}, \bibinfo{person}{Ioannis Kangas},
  \bibinfo{person}{Maud Schwoerer}, \bibinfo{person}{Alessandro Mozzato},
  \bibinfo{person}{Michele Ferretti}, \bibinfo{person}{Korvesis Panagiotis},
  {and} \bibinfo{person}{Lucas Bernardi}.} \bibinfo{year}{2021}\natexlab{b}.
\newblock \showarticletitle{I Know What You Did Next Summer: Challenges in
  Travel Destination Recommendation}.
\newblock  (\bibinfo{year}{2021}).
\newblock


\bibitem[\protect\citeauthoryear{Hong, Kveton, Zaheer, Chow, and Ahmed}{Hong
  et~al\mbox{.}}{2021}]%
        {hong2021non}
\bibfield{author}{\bibinfo{person}{Joey Hong}, \bibinfo{person}{Branislav
  Kveton}, \bibinfo{person}{Manzil Zaheer}, \bibinfo{person}{Yinlam Chow},
  {and} \bibinfo{person}{Amr Ahmed}.} \bibinfo{year}{2021}\natexlab{}.
\newblock \showarticletitle{Non-Stationary Off-Policy Optimization}. In
  \bibinfo{booktitle}{\emph{International Conference on Artificial Intelligence
  and Statistics}}. PMLR, \bibinfo{pages}{2494--2502}.
\newblock


\bibitem[\protect\citeauthoryear{Huang, Oosterhuis, de~Rijke, and van
  Hoof}{Huang et~al\mbox{.}}{2020}]%
        {huang2020keeping}
\bibfield{author}{\bibinfo{person}{Jin Huang}, \bibinfo{person}{Harrie
  Oosterhuis}, \bibinfo{person}{Maarten de Rijke}, {and} \bibinfo{person}{Herke
  van Hoof}.} \bibinfo{year}{2020}\natexlab{}.
\newblock \showarticletitle{Keeping Dataset Biases out of the Simulation: A
  Debiased Simulator for Reinforcement Learning based Recommender Systems}. In
  \bibinfo{booktitle}{\emph{Fourteenth ACM Conference on Recommender Systems}}.
  \bibinfo{pages}{190--199}.
\newblock


\bibitem[\protect\citeauthoryear{Ie, wei Hsu, Mladenov, Jain, Narvekar, Wang,
  Wu, and Boutilier}{Ie et~al\mbox{.}}{2019}]%
        {ie2019recsim}
\bibfield{author}{\bibinfo{person}{Eugene Ie}, \bibinfo{person}{Chih wei Hsu},
  \bibinfo{person}{Martin Mladenov}, \bibinfo{person}{Vihan Jain},
  \bibinfo{person}{Sanmit Narvekar}, \bibinfo{person}{Jing Wang},
  \bibinfo{person}{Rui Wu}, {and} \bibinfo{person}{Craig Boutilier}.}
  \bibinfo{year}{2019}\natexlab{}.
\newblock \showarticletitle{RecSim: A Configurable Simulation Platform for
  Recommender Systems}.
\newblock  (\bibinfo{year}{2019}).
\newblock
\showeprint[arxiv]{1909.04847}~[cs.LG]


\bibitem[\protect\citeauthoryear{Jagerman, Oosterhuis, and de~Rijke}{Jagerman
  et~al\mbox{.}}{2019}]%
        {jagerman2019model}
\bibfield{author}{\bibinfo{person}{Rolf Jagerman}, \bibinfo{person}{Harrie
  Oosterhuis}, {and} \bibinfo{person}{Maarten de Rijke}.}
  \bibinfo{year}{2019}\natexlab{}.
\newblock \showarticletitle{To model or to intervene: A comparison of
  counterfactual and online learning to rank from user interactions}. In
  \bibinfo{booktitle}{\emph{Proceedings of the 42nd international ACM SIGIR
  conference on research and development in information retrieval}}.
  \bibinfo{pages}{15--24}.
\newblock


\bibitem[\protect\citeauthoryear{Joulani, Gyorgy, and Szepesv{\'a}ri}{Joulani
  et~al\mbox{.}}{2013}]%
        {joulani2013online}
\bibfield{author}{\bibinfo{person}{Pooria Joulani}, \bibinfo{person}{Andras
  Gyorgy}, {and} \bibinfo{person}{Csaba Szepesv{\'a}ri}.}
  \bibinfo{year}{2013}\natexlab{}.
\newblock \showarticletitle{Online learning under delayed feedback}. In
  \bibinfo{booktitle}{\emph{International Conference on Machine Learning}}.
  PMLR, \bibinfo{pages}{1453--1461}.
\newblock


\bibitem[\protect\citeauthoryear{Kleinberg, Niculescu-Mizil, and
  Sharma}{Kleinberg et~al\mbox{.}}{2010}]%
        {kleinberg2010regret}
\bibfield{author}{\bibinfo{person}{Robert Kleinberg},
  \bibinfo{person}{Alexandru Niculescu-Mizil}, {and} \bibinfo{person}{Yogeshwer
  Sharma}.} \bibinfo{year}{2010}\natexlab{}.
\newblock \showarticletitle{Regret bounds for sleeping experts and bandits}.
\newblock \bibinfo{journal}{\emph{Machine learning}} \bibinfo{volume}{80},
  \bibinfo{number}{2} (\bibinfo{year}{2010}), \bibinfo{pages}{245--272}.
\newblock


\bibitem[\protect\citeauthoryear{Ktena, Tejani, Theis, Myana, Dilipkumar,
  Husz{\'a}r, Yoo, and Shi}{Ktena et~al\mbox{.}}{2019}]%
        {ktena2019addressing}
\bibfield{author}{\bibinfo{person}{Sofia~Ira Ktena}, \bibinfo{person}{Alykhan
  Tejani}, \bibinfo{person}{Lucas Theis}, \bibinfo{person}{Pranay~Kumar Myana},
  \bibinfo{person}{Deepak Dilipkumar}, \bibinfo{person}{Ferenc Husz{\'a}r},
  \bibinfo{person}{Steven Yoo}, {and} \bibinfo{person}{Wenzhe Shi}.}
  \bibinfo{year}{2019}\natexlab{}.
\newblock \showarticletitle{Addressing delayed feedback for continuous training
  with neural networks in CTR prediction}. In
  \bibinfo{booktitle}{\emph{Proceedings of the 13th ACM conference on
  recommender systems}}. \bibinfo{pages}{187--195}.
\newblock


\bibitem[\protect\citeauthoryear{Lancewicki, Segal, Koren, and
  Mansour}{Lancewicki et~al\mbox{.}}{2021}]%
        {lancewicki2021stochastic}
\bibfield{author}{\bibinfo{person}{Tal Lancewicki}, \bibinfo{person}{Shahar
  Segal}, \bibinfo{person}{Tomer Koren}, {and} \bibinfo{person}{Yishay
  Mansour}.} \bibinfo{year}{2021}\natexlab{}.
\newblock \showarticletitle{Stochastic multi-armed bandits with unrestricted
  delay distributions}. In \bibinfo{booktitle}{\emph{International Conference
  on Machine Learning}}. PMLR, \bibinfo{pages}{5969--5978}.
\newblock


\bibitem[\protect\citeauthoryear{Lattimore and Szepesv{\'a}ri}{Lattimore and
  Szepesv{\'a}ri}{2020}]%
        {lattimore2020bandit}
\bibfield{author}{\bibinfo{person}{Tor Lattimore} {and} \bibinfo{person}{Csaba
  Szepesv{\'a}ri}.} \bibinfo{year}{2020}\natexlab{}.
\newblock \bibinfo{booktitle}{\emph{Bandit algorithms}}.
\newblock \bibinfo{publisher}{Cambridge University Press}.
\newblock


\bibitem[\protect\citeauthoryear{Liu and Ho}{Liu and Ho}{2018}]%
        {liu2018incentivizing}
\bibfield{author}{\bibinfo{person}{Yang Liu} {and} \bibinfo{person}{Chien-Ju
  Ho}.} \bibinfo{year}{2018}\natexlab{}.
\newblock \showarticletitle{Incentivizing high quality user contributions: New
  arm generation in bandit learning}. In \bibinfo{booktitle}{\emph{Proceedings
  of the AAAI Conference on Artificial Intelligence}},
  Vol.~\bibinfo{volume}{32}.
\newblock


\bibitem[\protect\citeauthoryear{Liu and Li}{Liu and Li}{2021}]%
        {liu2021map}
\bibfield{author}{\bibinfo{person}{Yi Liu} {and} \bibinfo{person}{Lihong Li}.}
  \bibinfo{year}{2021}\natexlab{}.
\newblock \showarticletitle{A map of bandits for e-commerce}. In
  \bibinfo{booktitle}{\emph{KDD 2021 Workshop on Multi-Armed Bandits and
  Reinforcement Learning (MARBLE)}}.
\newblock
\urldef\tempurl%
\url{https://www.amazon.science/publications/a-map-of-bandits-for-e-commerce}
\showURL{%
\tempurl}


\bibitem[\protect\citeauthoryear{Lu, Liu, Dong, Gu, Gama, and Zhang}{Lu
  et~al\mbox{.}}{2018}]%
        {lu2018learning}
\bibfield{author}{\bibinfo{person}{Jie Lu}, \bibinfo{person}{Anjin Liu},
  \bibinfo{person}{Fan Dong}, \bibinfo{person}{Feng Gu}, \bibinfo{person}{Joao
  Gama}, {and} \bibinfo{person}{Guangquan Zhang}.}
  \bibinfo{year}{2018}\natexlab{}.
\newblock \showarticletitle{Learning under concept drift: A review}.
\newblock \bibinfo{journal}{\emph{IEEE Transactions on Knowledge and Data
  Engineering}} \bibinfo{volume}{31}, \bibinfo{number}{12}
  (\bibinfo{year}{2018}), \bibinfo{pages}{2346--2363}.
\newblock


\bibitem[\protect\citeauthoryear{Luo, Wei, Agarwal, and Langford}{Luo
  et~al\mbox{.}}{2018}]%
        {luo2018efficient}
\bibfield{author}{\bibinfo{person}{Haipeng Luo}, \bibinfo{person}{Chen-Yu Wei},
  \bibinfo{person}{Alekh Agarwal}, {and} \bibinfo{person}{John Langford}.}
  \bibinfo{year}{2018}\natexlab{}.
\newblock \showarticletitle{Efficient contextual bandits in non-stationary
  worlds}. In \bibinfo{booktitle}{\emph{Conference On Learning Theory}}. PMLR,
  \bibinfo{pages}{1739--1776}.
\newblock


\bibitem[\protect\citeauthoryear{Mehrotra, Xue, and Lalmas}{Mehrotra
  et~al\mbox{.}}{2020}]%
        {10.1145/3394486.3403374}
\bibfield{author}{\bibinfo{person}{Rishabh Mehrotra}, \bibinfo{person}{Niannan
  Xue}, {and} \bibinfo{person}{Mounia Lalmas}.}
  \bibinfo{year}{2020}\natexlab{}.
\newblock \showarticletitle{Bandit Based Optimization of Multiple Objectives on
  a Music Streaming Platform}. In \bibinfo{booktitle}{\emph{Proceedings of the
  26th ACM SIGKDD International Conference on Knowledge Discovery \&amp; Data
  Mining}} (Virtual Event, CA, USA) \emph{(\bibinfo{series}{KDD '20})}.
  \bibinfo{publisher}{Association for Computing Machinery},
  \bibinfo{address}{New York, NY, USA}, \bibinfo{pages}{3224–3233}.
\newblock
\showISBNx{9781450379984}
\urldef\tempurl%
\url{https://doi.org/10.1145/3394486.3403374}
\showDOI{\tempurl}


\bibitem[\protect\citeauthoryear{Nabi, Nassif, Hong, Mamani, and Imbens}{Nabi
  et~al\mbox{.}}{2021}]%
        {Nabi2021}
\bibfield{author}{\bibinfo{person}{Sareh Nabi}, \bibinfo{person}{Houssam
  Nassif}, \bibinfo{person}{Joseph Hong}, \bibinfo{person}{Hamed Mamani}, {and}
  \bibinfo{person}{Guido Imbens}.} \bibinfo{year}{2021}\natexlab{}.
\newblock \showarticletitle{Bayesian meta-prior learning using Empirical
  Bayes}.
\newblock \bibinfo{journal}{\emph{The Journal of Management Science}}
  (\bibinfo{year}{2021}).
\newblock
\urldef\tempurl%
\url{https://www.amazon.science/publications/bayesian-meta-prior-learning-using-empirical-bayes}
\showURL{%
\tempurl}


\bibitem[\protect\citeauthoryear{Oosterhuis and de~Rijke}{Oosterhuis and
  de~Rijke}{2021}]%
        {oosterhuis2021unifying}
\bibfield{author}{\bibinfo{person}{Harrie Oosterhuis} {and}
  \bibinfo{person}{Maarten de Rijke}.} \bibinfo{year}{2021}\natexlab{}.
\newblock \showarticletitle{Unifying online and counterfactual learning to
  rank: A novel counterfactual estimator that effectively utilizes online
  interventions}. In \bibinfo{booktitle}{\emph{Proceedings of the 14th ACM
  International Conference on Web Search and Data Mining}}.
  \bibinfo{pages}{463--471}.
\newblock


\bibitem[\protect\citeauthoryear{O’Donoghue, Munos, Kavukcuoglu, and
  Mnih}{O’Donoghue et~al\mbox{.}}{2016}]%
        {o2016pgq}
\bibfield{author}{\bibinfo{person}{Brendan O’Donoghue},
  \bibinfo{person}{R{\'e}mi Munos}, \bibinfo{person}{Koray Kavukcuoglu}, {and}
  \bibinfo{person}{Volodymyr Mnih}.} \bibinfo{year}{2016}\natexlab{}.
\newblock \showarticletitle{PGQ: Combining policy gradient and Q-learning}.
\newblock \bibinfo{journal}{\emph{ICML}} (\bibinfo{year}{2016}).
\newblock


\bibitem[\protect\citeauthoryear{Rohde, Bonner, Dunlop, Vasile, and
  Karatzoglou}{Rohde et~al\mbox{.}}{2018}]%
        {rohde2018recogym}
\bibfield{author}{\bibinfo{person}{David Rohde}, \bibinfo{person}{Stephen
  Bonner}, \bibinfo{person}{Travis Dunlop}, \bibinfo{person}{Flavian Vasile},
  {and} \bibinfo{person}{Alexandros Karatzoglou}.}
  \bibinfo{year}{2018}\natexlab{}.
\newblock \showarticletitle{RecoGym: A Reinforcement Learning Environment for
  the problem of Product Recommendation in Online Advertising}.
\newblock \bibinfo{journal}{\emph{arXiv preprint arXiv:1808.00720}}
  (\bibinfo{year}{2018}).
\newblock


\bibitem[\protect\citeauthoryear{Saha and Gupta}{Saha and Gupta}{2022}]%
        {saha2022non-stationary}
\bibfield{author}{\bibinfo{person}{Aadirupa Saha} {and}
  \bibinfo{person}{Shubham Gupta}.} \bibinfo{year}{2022}\natexlab{}.
\newblock \showarticletitle{Non-Stationary Dueling Bandits}. In
  \bibinfo{booktitle}{\emph{ICML 2022}}.
\newblock
\urldef\tempurl%
\url{https://www.microsoft.com/en-us/research/publication/non-stationary-dueling-bandits/}
\showURL{%
\tempurl}


\bibitem[\protect\citeauthoryear{Saito, Shunsuke, Megumi, and Yusuke}{Saito
  et~al\mbox{.}}{2020}]%
        {saito2020open}
\bibfield{author}{\bibinfo{person}{Yuta Saito}, \bibinfo{person}{Aihara
  Shunsuke}, \bibinfo{person}{Matsutani Megumi}, {and} \bibinfo{person}{Narita
  Yusuke}.} \bibinfo{year}{2020}\natexlab{}.
\newblock \showarticletitle{Open Bandit Dataset and Pipeline: Towards Realistic
  and Reproducible Off-Policy Evaluation}.
\newblock \bibinfo{journal}{\emph{arXiv preprint arXiv:2008.07146}}
  (\bibinfo{year}{2020}).
\newblock


\bibitem[\protect\citeauthoryear{Sajeev, Huang, Karampatziakis, Hall, Kochman,
  and Chen}{Sajeev et~al\mbox{.}}{2021}]%
        {10.1145/3447548.3467165}
\bibfield{author}{\bibinfo{person}{Sandra Sajeev}, \bibinfo{person}{Jade
  Huang}, \bibinfo{person}{Nikos Karampatziakis}, \bibinfo{person}{Matthew
  Hall}, \bibinfo{person}{Sebastian Kochman}, {and} \bibinfo{person}{Weizhu
  Chen}.} \bibinfo{year}{2021}\natexlab{}.
\newblock \showarticletitle{Contextual Bandit Applications in a Customer
  Support Bot}. In \bibinfo{booktitle}{\emph{Proceedings of the 27th ACM SIGKDD
  Conference on Knowledge Discovery \& Data Mining}} (Virtual Event, Singapore)
  \emph{(\bibinfo{series}{KDD '21})}. \bibinfo{publisher}{Association for
  Computing Machinery}, \bibinfo{address}{New York, NY, USA},
  \bibinfo{pages}{3522–3530}.
\newblock
\showISBNx{9781450383325}
\urldef\tempurl%
\url{https://doi.org/10.1145/3447548.3467165}
\showDOI{\tempurl}


\bibitem[\protect\citeauthoryear{Shi, Yu, Da, Chen, and Zeng}{Shi
  et~al\mbox{.}}{2019}]%
        {shi2019virtual}
\bibfield{author}{\bibinfo{person}{Jing-Cheng Shi}, \bibinfo{person}{Yang Yu},
  \bibinfo{person}{Qing Da}, \bibinfo{person}{Shi-Yong Chen}, {and}
  \bibinfo{person}{An-Xiang Zeng}.} \bibinfo{year}{2019}\natexlab{}.
\newblock \showarticletitle{Virtual-taobao: Virtualizing real-world online
  retail environment for reinforcement learning}. In
  \bibinfo{booktitle}{\emph{Proceedings of the AAAI Conference on Artificial
  Intelligence}}, Vol.~\bibinfo{volume}{33}. \bibinfo{pages}{4902--4909}.
\newblock


\bibitem[\protect\citeauthoryear{Slivkins and Upfal}{Slivkins and
  Upfal}{2008}]%
        {slivkins2008adapting}
\bibfield{author}{\bibinfo{person}{Aleksandrs Slivkins} {and}
  \bibinfo{person}{Eli Upfal}.} \bibinfo{year}{2008}\natexlab{}.
\newblock \showarticletitle{Adapting to a Changing Environment: the Brownian
  Restless Bandits.}. In \bibinfo{booktitle}{\emph{COLT}}.
  \bibinfo{pages}{343--354}.
\newblock


\bibitem[\protect\citeauthoryear{Strehl, Langford, Li, and Kakade}{Strehl
  et~al\mbox{.}}{2010}]%
        {strehl2010learning}
\bibfield{author}{\bibinfo{person}{Alex Strehl}, \bibinfo{person}{John
  Langford}, \bibinfo{person}{Lihong Li}, {and} \bibinfo{person}{Sham~M
  Kakade}.} \bibinfo{year}{2010}\natexlab{}.
\newblock \showarticletitle{Learning from logged implicit exploration data}.
\newblock \bibinfo{journal}{\emph{Advances in neural information processing
  systems}}  \bibinfo{volume}{23} (\bibinfo{year}{2010}).
\newblock


\bibitem[\protect\citeauthoryear{Tax, Vries, Jong, Dosoula, Akker, Smith,
  Thuong, and Bernardi}{Tax et~al\mbox{.}}{2021}]%
        {tax2021machine}
\bibfield{author}{\bibinfo{person}{Niek Tax}, \bibinfo{person}{Kees Jan~de
  Vries}, \bibinfo{person}{Mathijs~de Jong}, \bibinfo{person}{Nikoleta
  Dosoula}, \bibinfo{person}{Bram van~den Akker}, \bibinfo{person}{Jon Smith},
  \bibinfo{person}{Olivier Thuong}, {and} \bibinfo{person}{Lucas Bernardi}.}
  \bibinfo{year}{2021}\natexlab{}.
\newblock \showarticletitle{Machine learning for fraud detection in e-Commerce:
  A research agenda}. In \bibinfo{booktitle}{\emph{International Workshop on
  Deployable Machine Learning for Security Defense}}. Springer,
  \bibinfo{pages}{30--54}.
\newblock


\bibitem[\protect\citeauthoryear{Thune, Cesa-Bianchi, and Seldin}{Thune
  et~al\mbox{.}}{2019}]%
        {thune2019nonstochastic}
\bibfield{author}{\bibinfo{person}{Tobias~Sommer Thune},
  \bibinfo{person}{Nicol{\`o} Cesa-Bianchi}, {and} \bibinfo{person}{Yevgeny
  Seldin}.} \bibinfo{year}{2019}\natexlab{}.
\newblock \showarticletitle{Nonstochastic multiarmed bandits with unrestricted
  delays}.
\newblock \bibinfo{journal}{\emph{Advances in Neural Information Processing
  Systems}}  \bibinfo{volume}{32} (\bibinfo{year}{2019}).
\newblock


\bibitem[\protect\citeauthoryear{Tsagkias, King, Kallumadi, Murdock, and
  de~Rijke}{Tsagkias et~al\mbox{.}}{2021}]%
        {tsagkias2021challenges}
\bibfield{author}{\bibinfo{person}{Manos Tsagkias},
  \bibinfo{person}{Tracy~Holloway King}, \bibinfo{person}{Surya Kallumadi},
  \bibinfo{person}{Vanessa Murdock}, {and} \bibinfo{person}{Maarten de Rijke}.}
  \bibinfo{year}{2021}\natexlab{}.
\newblock \showarticletitle{Challenges and research opportunities in ecommerce
  search and recommendations}. In \bibinfo{booktitle}{\emph{ACM SIGIR Forum}},
  Vol.~\bibinfo{volume}{54}. ACM New York, NY, USA, \bibinfo{pages}{1--23}.
\newblock


\bibitem[\protect\citeauthoryear{Vamsi~Potluru and Veloso}{Vamsi~Potluru and
  Veloso}{2021}]%
        {liu2021funnel}
\bibfield{author}{\bibinfo{person}{Sameena~Shah Vamsi~Potluru,
  Branislav~Kveton} {and} \bibinfo{person}{Manuela Veloso}.}
  \bibinfo{year}{2021}\natexlab{}.
\newblock \showarticletitle{Funnel Bandits}. In \bibinfo{booktitle}{\emph{KDD
  2021 Workshop on Multi-Armed Bandits and Reinforcement Learning (MARBLE)}}.
\newblock


\bibitem[\protect\citeauthoryear{Zeng, Wang, Mokhtari, and Li}{Zeng
  et~al\mbox{.}}{2016}]%
        {zeng2016online}
\bibfield{author}{\bibinfo{person}{Chunqiu Zeng}, \bibinfo{person}{Qing Wang},
  \bibinfo{person}{Shekoofeh Mokhtari}, {and} \bibinfo{person}{Tao Li}.}
  \bibinfo{year}{2016}\natexlab{}.
\newblock \showarticletitle{Online context-aware recommendation with time
  varying multi-armed bandit}. In \bibinfo{booktitle}{\emph{Proceedings of the
  22nd ACM SIGKDD international conference on Knowledge discovery and data
  mining}}. \bibinfo{pages}{2025--2034}.
\newblock


\bibitem[\protect\citeauthoryear{{\v{Z}}liobait{\.e}, Pechenizkiy, and
  Gama}{{\v{Z}}liobait{\.e} et~al\mbox{.}}{2016}]%
        {vzliobaite2016overview}
\bibfield{author}{\bibinfo{person}{Indr{\.e} {\v{Z}}liobait{\.e}},
  \bibinfo{person}{Mykola Pechenizkiy}, {and} \bibinfo{person}{Joao Gama}.}
  \bibinfo{year}{2016}\natexlab{}.
\newblock \showarticletitle{An overview of concept drift applications}.
\newblock \bibinfo{journal}{\emph{Big data analysis: new algorithms for a new
  society}} (\bibinfo{year}{2016}), \bibinfo{pages}{91--114}.
\newblock


\bibitem[\protect\citeauthoryear{Zuo}{Zuo}{2021}]%
        {zuo2021recsys}
\bibfield{author}{\bibinfo{person}{Xingdong Zuo}.}
  \bibinfo{year}{2021}\natexlab{}.
\newblock \bibinfo{booktitle}{\emph{gym-recsys: Customizable RecSys Simulator
  for OpenAI Gym}}.
\newblock
\urldef\tempurl%
\url{https://github.com/zuoxingdong/gym-recsys}
\showURL{%
\tempurl}


\end{thebibliography}
\pagebreak

\begin{appendices}
\section{Example Experiments}
To provide future practitioners with a starting point to use our simulations, we highlight some examples of experiments in this appendix. Each experiment demonstrates how a mentioned challenge can affect the learning of various bandit algorithms and off-policy methods. 

\subsection{Off-Policy Outperforming On-Policy} \label{sec:offpolicyappendix}
In this appendix we compare the performance of various standard bandit algorithms with an off-policy learned policy using Inverse Propensity Weighting (IPW). Each bandit is first optimised for \textit{n} rounds, after which we show the performance on an evaluation period of the same number of rounds. The IPW learning is trained on the logged data of a training period of \textit{n} rounds and then also evaluated on the same dataset as the bandits. In contrast to IPW, all the bandit models can continue learning during the evaluation period. Each policy needs to optimise a stationary environment with 10 arms. 

\begin{figure}[h]
\centering
  \centering
  \includegraphics[width=.8\linewidth, trim={-4cm 0 0 0cm}]{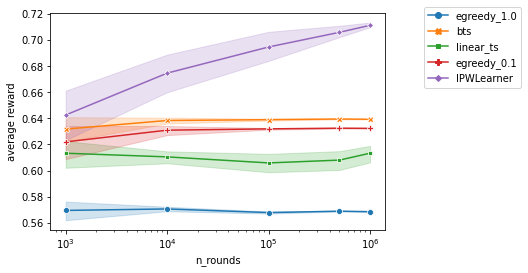}
  \caption{A comparison of various on-policy and off-policy methods when different amounts of rounds are available for optimisation and evaluation.}
\label{fig:onoff}
\end{figure}

In figure~\ref{fig:onoff}, we clearly see that the IPW learner starts outperforming the bandit policies after approximately 5,000 rounds. This aligns with some of our own experiments where an IPW learner showed significantly better performance than bandit algorithms that have been running on-policy. Practitioners can adjust use similar setup's to understand how to answer \ref{scenariooffpolicy} for their own situation.

\begin{figure}[h]
\centering
  \centering
  \includegraphics[width=1.0\linewidth, trim={-7cm 0 0 0}]{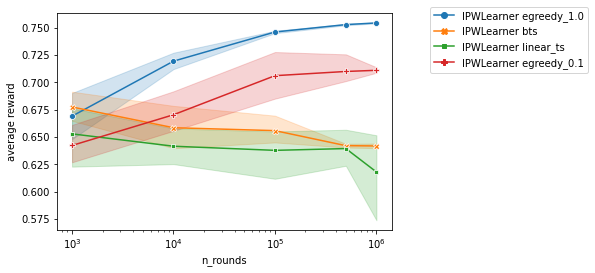}
  \caption{An IPW learner trained on logs created by various bandit algorithms.}
\label{fig:offdiffpolicies}
\end{figure}

In Figure~\ref{fig:onoff}, the IPW learner is trained on the logs produced by the \textit{egreedy\_0.1} model with 10\% random exploration (epsilon) traffic. However, using the simulations we can also easily compare how the IPW learner would perform using logs created by different bandit policies (\ref{strategycounterfactual}). In Figure~\ref{fig:offdiffpolicies}, we show how the performance of the IPW learner can significantly differ depending on the logging policy. This difference in performance is likely caused by the quality of the learned propensity model feeding into the IPW learner.

\subsection{Drift and On-Policy Bandits}  \label{sec:driftappendix}
In this appendix, we demonstrate how the effect of drift of bandit algorithms (\ref{question:drift}) can be observed using the simulation described in this paper. 
Two types of simulations are shown: 1) Sudden drift after 25,000 steps, and 2) dual seasonal drift shifting every 5000 steps for a total of 50,000 steps, i.e.\ 5 full seasons of each type.

In both simulations, we run Epsilon Greedy (0.10 epsilon), binary Thompson sampling (bts), Linucb, and Linear Thompson (lints) sampling. Additionally, we show how bts and a fully random policy would have performed on non-drifting data. During the drift the base coefficient weight is set to 0.3 to sustain some parts of the original coefficients while drifting.

\begin{figure}[h]
\centering
\begin{subfigure}{.4\textwidth}
  \centering
  \includegraphics[width=\linewidth, trim={0 0 5.1cm 0},clip]{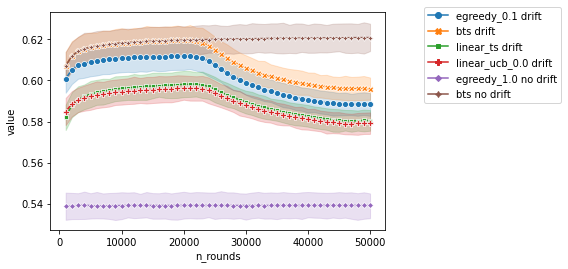}
  \caption{Abrupt drift starting at step 20,000 and fully shifted at step 25,000, the base coefficient is weighted at .30.}
  \label{fig:sub1}
\end{subfigure}%
\begin{subfigure}{.15\textwidth}
  \centering
  \includegraphics[width=\linewidth, trim={14.7cm 0 0 0},clip]{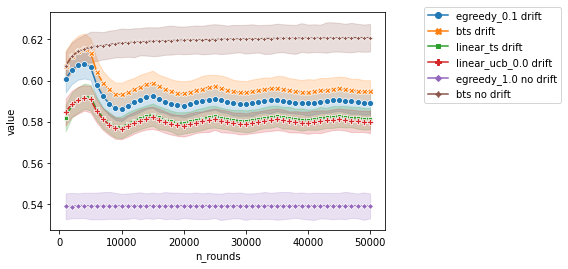}
  \label{fig:legend}
\end{subfigure}
\begin{subfigure}{.4\textwidth}
  \centering
  \includegraphics[width=\linewidth, trim={0 0 5.1cm 0},clip]{images/seasonal_drift.png}
  \caption{Seasonal drift between two seasons every 5,000 steps. The base coefficient is weighted at .30.}
  \label{fig:sub2}
\end{subfigure}
\caption{Comparing two types of drift with the performance of bts and random policies in a stationary environment. }
\label{fig:drift}
\end{figure}

In Figure~\ref{fig:drift} we see that all the bandit algorithms lose significant reward when the environment starts shifting. Neither of the experiments show that the bandit algorithms are able to recover the lost reward within a reasonable timeframe. Here we call "reasonable" a similar timeframe it took to get to the original performance of the bandit

Moreover, we can observe that the seasonal bandit stabilizes at the end of the simulation at a subpar performance compared to the stationary bandit.

\subsection{Unbiased and Reward-Depend Delay}  \label{sec:rewardappendix}
In this appendix we demonstrate how to simulate different effects of delayed reward on the learning ability of bandit policies.

\begin{figure}[h]
\centering
  \centering
  \includegraphics[width=.5\linewidth]{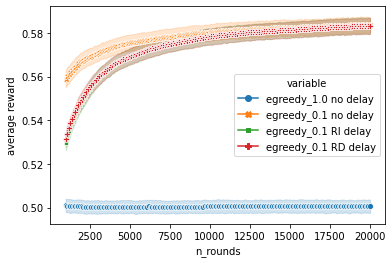}
  \caption{A comparison of count-based epsilon greedy under drift in a stationary environment.}
\label{fig:delay}
\end{figure}

In Figure~\ref{fig:delay} we compare a count-based epsilon greedy policy with 10\% exploration under unbiased exponential delay, reward-dependent exponential delay, and no delay (\ref{question:delaybias}). We've fixed the exponential scale between 900, for the reward-dependent experiment, and 1000 and ran the experiment 1000 times for statistical power. It is apparent from the figure that both policies optimised under delay take longer to converge. However, after approximately 10,000 steps both the delayed and non-delayed policies are receiving an non-significant different average reward. This plot seems to indicate that the bandit policy can be optimised under delayed reward with minimal cost to the total reward.

\begin{figure}[h]
\centering
  \centering
  \includegraphics[width=1.0\linewidth, trim={-8cm 0 0 0}]{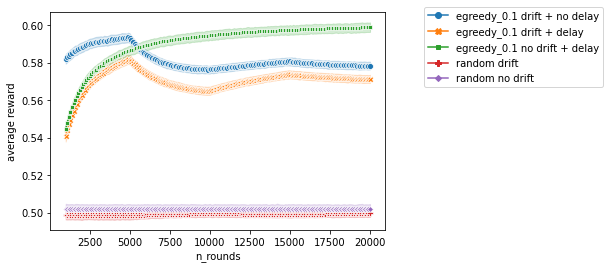}
  \caption{A comparison of count-based epsilon greedy under drift in a non-stationary (seasonal) environment.}
\label{fig:delaynonstationary}
\end{figure}

However, when we introduce a non-stationary seasonal environment in figure~\ref{fig:delaynonstationary}, we observe that the drift significantly affects the total reward. 

\end{appendices}


\end{document}